\documentclass[10pt,twocolumn,letterpaper]{article}

\usepackage{cvpr}
\usepackage{times}
\usepackage{epsfig}
\usepackage{graphicx}

\usepackage{amsmath,amssymb,amsfonts,dsfont,bm,mathrsfs}
\usepackage{xcolor}
\usepackage{caption}
\usepackage{booktabs,multirow,adjustbox,multicol}
\usepackage{float}
\usepackage[pagebackref=true,breaklinks=true,letterpaper=true,colorlinks,bookmarks=false]{hyperref}
\newcommand{\R}{\mathbb{R}}       % notation of real number space.
\newcommand{\z}{{\rm\bf z}}       % notation of latent code.
\newcommand{\Z}{\mathcal{Z}}      % notation of latent space.
\newcommand{\x}{{\rm\bf x}}       % notation of image.
\newcommand{\X}{\mathcal{X}}      % notation of image space.
\newcommand{\F}{{\rm\bf F}}       % notation of intermediate feature.
\newcommand{\Loss}{\mathcal{L}}   % notation of loss function.

\def\cvprPaperID{1107}
\cvprfinalcopy
\begin{document}

%%%% Title
\title{Image Processing Using Multi-Code GAN Prior}
\author{
  Jinjin Gu\textsuperscript{1,2},
  Yujun Shen\textsuperscript{1},
  Bolei Zhou\textsuperscript{1} \\
   \textsuperscript{1}The Chinese University of Hong Kong \quad
  \textsuperscript{2}The Chinese University of Hong Kong, Shenzhen \\
  {\tt\small
    jinjingu@link.cuhk.edu.cn,
    \{sy116, bzhou\}@ie.cuhk.edu.hk
  }
}

%%%% Figure: Teaser
\twocolumn[{
\renewcommand\twocolumn[1][]{#1}
\maketitle
\begin{center}
  \vspace{-5pt}
  \includegraphics[width=0.95\linewidth]{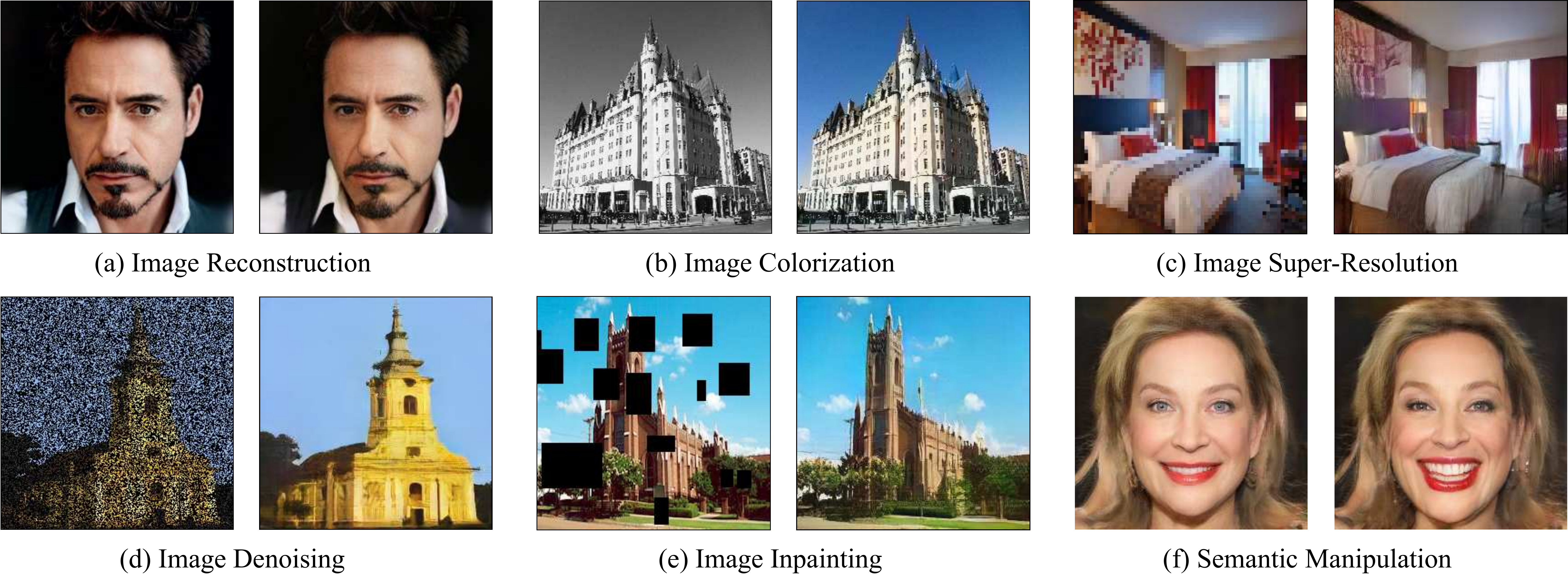}
  \vspace{-8pt}
  \captionsetup{type=figure,font=small}
  \caption{
    Multi-code GAN prior facilitates many image processing applications using the reconstruction from \emph{fixed} PGGAN \cite{pggan} models.
  }
  \label{fig:teaser}
  \vspace{2pt}
\end{center}
}]

% \thispagestyle{empty}

%%%% Abstract
\begin{abstract}
%%%
Despite the success of Generative Adversarial Networks (GANs) in image synthesis, applying trained GAN models to real image processing remains challenging.
Previous methods typically invert a target image back to the latent space either by back-propagation or by learning an additional encoder.
However, the reconstructions from both of the methods are far from ideal.
In this work, we propose a novel approach, called mGANprior, to incorporate the well-trained GANs as effective prior to a variety of image processing tasks.
In particular, we employ multiple latent codes to generate multiple feature maps at some intermediate layer of the generator, then compose them with adaptive channel importance to recover the input image.
Such an over-parameterization of the latent space significantly improves the image reconstruction quality, outperforming existing competitors.
The resulting high-fidelity image reconstruction enables the trained GAN models as prior to many real-world applications, such as image colorization, super-resolution, image inpainting, and semantic manipulation.
We further analyze the properties of the layer-wise representation learned by GAN models and shed light on what knowledge each layer is capable of representing.\footnote{Code is available at \href{https://genforce.github.io/mganprior/}{this link}.}
\end{abstract}

%%%% Section: Introduction
\vspace{-12pt}
\section{Introduction}\label{sec:introduction}
%%%%
Recently, Generative Adversarial Networks (GANs) \cite{gan} have advanced image generation by improving the synthesis quality \cite{pggan,biggan,stylegan} and stabilizing the training process \cite{wgan,began,wgangp}.
The capability to produce high-quality images makes GANs applicable to many image processing tasks, such as semantic face editing \cite{lample2017fader,shen2018faceid}, super-resolution \cite{ledig2017photo,wang2018esrgan}, image-to-image translation \cite{zhu2017unpaired,stargan,liu2019few}, \emph{etc}.
However, most of these GAN-based approaches require special design of network structures \cite{lample2017fader,zhu2017unpaired} or loss functions \cite{shen2018faceid,ledig2017photo} for a particular task, limiting their generalization ability.
On the other hand, the large-scale GAN models, like StyleGAN \cite{stylegan} and BigGAN \cite{biggan}, can synthesize photo-realistic images after being trained with millions of diverse images.
Their neural representations are shown to contain various levels of semantics underlying the observed data \cite{jahanian2020steerability,goetschalckx2019ganalyze,shen2020interpreting,yang2019semantic}.
Reusing these models as prior to real image processing with minor effort could potentially lead to wider applications but remains much less explored.

The main challenge towards this goal is that the standard GAN model is initially designed for synthesizing images from random noises, thus is unable to take real images for any post-processing.
A common practice is to invert a given image back to a latent code such that it can be reconstructed by the generator.
In this way, the inverted code can be used for further processing.
To reverse the generation process, existing approaches fall into two types.
One is to directly optimize the latent code by minimizing the reconstruction error through back-propagation \cite{lipton2017precise,inverting2018,invertibility}.
The other is to train an extra encoder to learn the mapping from the image space to the latent space \cite{inverting2016,zhu2016generative,bau2019seeing,inverting2019}.
However, the reconstructions achieved by both methods are far from ideal, especially when the given image is with high resolution.
Consequently, the reconstructed image with low quality is unable to be used for image processing tasks.

In principle, it is impossible to recover every detail of any arbitrary real image using a single latent code, otherwise, we would have an unbeatable image compression method.
In other words, the expressiveness of the latent code is limited due to its finite dimensionality.
%since the dimension of the latent space of a trained GAN model is fixed.
%
Therefore, to faithfully recover a target image, we propose to employ \emph{multiple} latent codes and compose their corresponding feature maps at some intermediate layer of the generator.
Utilizing multiple latent codes allows the generator to recover the target image using all the possible composition knowledge learned in the deep generative representation.
The experiments show that our approach significantly improves the image reconstruction quality.
More importantly, being able to better reconstruct the input image, our approach facilitates various real image processing applications by using pre-trained GAN models as prior \emph{without} retraining or modification, which is shown in Fig.\ref{fig:teaser}.
We summarize our contributions as follows:
\begin{itemize}
  \vspace{-5pt}
  \setlength{\itemsep}{0pt}
  \setlength{\parsep}{0pt}
  \setlength{\parskip}{0pt}
  \item We propose \emph{mGANprior}, shorted for multi-code GAN prior, as an effective GAN inversion method by using multiple latent codes and adaptive channel importance. The method faithfully reconstructs the given real image, surpassing existing approaches.
  \item We apply the proposed mGANprior to a range of real-world applications, such as image colorization, super-resolution, image inpainting, semantic manipulation, \emph{etc}, demonstrating its potential in real image processing.
  \item We further analyze the internal representation of different layers in a GAN generator by composing the features from the inverted latent codes at each layer respectively.
\end{itemize}

%%%% Section: Related Work
\section{Related Work}\label{sec:related-work}
%%%%

%%%% Related Work: GAN Inversion
\vspace{2pt}\noindent\textbf{GAN Inversion.}
The task of GAN inversion targets at reversing a given image back to a latent code with a pre-trained GAN model.
As an important step for applying GANs to real-world applications, it has attracted increasing attention recently.
To invert a fixed generator in GAN, existing methods either optimized the latent code based on gradient descent \cite{lipton2017precise,inverting2018,invertibility} or learned an extra encoder to project the image space back to the latent space \cite{inverting2016,zhu2016generative,bau2019seeing,inverting2019}.
Bau \emph{et al.} \cite{bau2019semantic} proposed to use encoder to provide better initialization for optimization.
There are also some models taking invertibility into account at the training stage \cite{dumoulin2016adversarially,donahue2016adversarial,glow}.
However, all the above methods only consider using a single latent code to recover the input image and the reconstruction quality is far from ideal, especially when the test image shows a huge domain gap to training data.
That is because the input image may not lie in the synthesis space of the generator, in which case the perfect inversion with a single latent code does not exist.
By contrast, we propose to increase the number of latent codes, which significantly improve the inversion quality no matter whether the target image is in-domain or out-of-domain.

%%%% Related Work: Image Processing with GANs
\vspace{2pt}\noindent\textbf{Image Processing with GANs.}
GANs have been widely used for real image processing due to its great power of synthesizing photo-realistic images.
These applications include image denoising \cite{chen2018image,kim2019grdn}, image inpainting \cite{yeh2017semantic,yu2018generative}, super-resolution \cite{ledig2017photo,wang2018esrgan}, image colorization \cite{suarez2017infrared,isola2017image}, style mixing \cite{hao2018mixgan,chen2018gated}, semantic image manipulation \cite{wang2018high,liang2018generative}, \emph{etc}.
However, current GAN-based models are usually designed for a particular task with specialized architectures \cite{hao2018mixgan,wang2018high} or loss functions \cite{ledig2017photo,chen2018gated}, and trained with paired data by taking one image as input and the other as supervision \cite{yeh2017semantic,isola2017image}.
Differently, our approach can reuse the knowledge contained in a well-trained GAN model and further enable a single GAN model as prior to all the aforementioned tasks \emph{without} retraining or modification.
It is worth noticing that our method can achieve similar or even better results than existing GAN-based methods that are particularly trained for a certain task.

%%%% Figure: Framework
\begin{figure*}[t]
  \centering
  \includegraphics[width=0.9\linewidth]{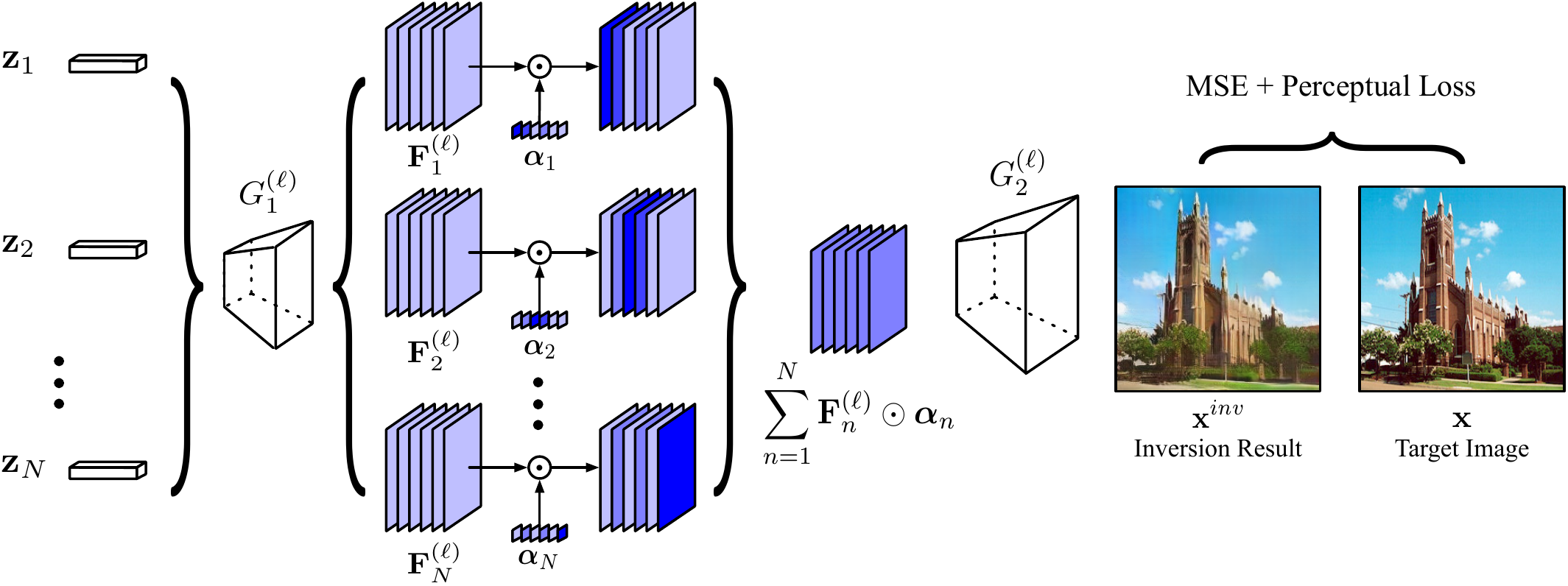}
  \vspace{-10pt}
  \captionsetup{font=small}
  \caption{
    Pipeline of GAN inversion using multiple latent codes $\{\z_n\}_{n=1}^N$.
    The generative features from these latent codes are composed at some intermediate layer (\emph{i.e.}, the $\ell$-th layer) of the generator, weighted by the adaptive channel importance scores $\{{\bm\alpha}_n\}_{n=1}^N$.
    All latent codes and the corresponding channel importance scores are jointly optimized to recover a target image.
  }
  \label{fig:framework}
 \vspace{-10pt}
\end{figure*}

%%%% Related Work: Deep Model Prior
\vspace{2pt}\noindent\textbf{Deep Model Prior.}
Generally, the impressive performance of the deep convolutional model can be attributed to its capacity of capturing statistical information from large-scale data as prior.
Such prior can be inversely used for image generation and image reconstruction \cite{upchurch2017deep,ulyanov2018deep,athar2019latent}.
Upchurch \emph{et al.} \cite{upchurch2017deep} inverted a discriminative model, starting from deep convolutional features, to achieve semantic image transformation.
Ulyanov \emph{et al.} \cite{ulyanov2018deep} reconstructed the target image with a U-Net structure to show that the structure of a generator network is sufficient to capture the low-level image statistics prior to any learning.
Athar \emph{et al.} \cite{athar2019latent} learned a universal image prior for a variety of image restoration tasks.
Some work theoretically explored the prior provided by deep generative models \cite{invertibility,hand2019global}, but the results using GAN prior to real image processing are still unsatisfying.
A recent work \cite{bau2019semantic} applied generative image prior to semantic photo manipulation, but it can only edit some partial regions of the input image yet fails to apply to other tasks like colorization or super-resolution.
That is because it only inverts the GAN model to some intermediate feature space instead of the earliest hidden space.
By contrast, our method reverses the entire generative process, \emph{i.e.}, from the image space to the initial latent space, which supports more flexible image processing tasks.

%%%% Section: Multi-Code GAN Prior
\section{Multi-Code GAN Prior}\label{sec:method}
%%%%
A well-trained generator $G(\cdot)$ of GAN can synthesize high-quality images by sampling codes from the latent space $\Z$.
Given a target image $\x$, the GAN inversion task aims at reversing the generation process by finding the adequate code to recover $\x$.
It can be formulated as
\begin{align}
  \z^* = \mathop{\arg\min}_{\z\in\Z} \Loss(G(\z), \x), \label{eq:single-code-inversion}
\end{align}
where $\Loss(\cdot,\cdot)$ denotes the objective function.

However, due to the highly non-convex natural of this optimization problem, previous methods fail to ideally reconstruct an arbitrary image by optimizing a single latent code.
To this end, we propose to use multiple latent codes and compose their corresponding intermediate feature maps with adaptive channel importance, as illustrated in Fig.\ref{fig:framework}.

%%%% Sub-Section: GAN Inversion with Multiple Latent Codes
\subsection{GAN Inversion with Multiple Latent Codes}\label{subsec:gan-inversion-with-multiple-latent-codes}
%%%%
The expressiveness of a single latent code may not be enough to recover all the details of a certain image.
Then, how about using $N$ latent codes $\{\z_n\}_{n=1}^N$, each of which can help reconstruct some sub-regions of the target image?
In the following, we introduce how to utilize multiple latent codes for GAN inversion.

%%%% GAN Inversion with Multiple Latent Codes: Feature Composition
\vspace{2pt}\noindent\textbf{Feature Composition.}
One key difficulty after introducing multiple latent codes is how to integrate them in the gene-ration process.
A straightforward solution is to fuse the images generated by each $\z_n$ from the image space $\X$.
However, $\X$ is not naturally a linear space such that linearly combining synthesized images is not guaranteed to produce a meaningful image, let alone recover the input in detail.
A recent work \cite{inverting2019} pointed out that inverting a generative model from the image space to some intermediate feature space is much easier than to the latent space.
Accordingly, we propose to combine the latent codes by composing their intermediate feature maps.
More concretely, the generator $G(\cdot)$ is divided into two sub-networks, \emph{i.e.}, $G_1^{(\ell)}(\cdot)$ and $G_2^{(\ell)}(\cdot)$.
Here, $\ell$ is the index of the intermediate layer to perform feature composition.
With such a separation, for any $\z_n$, we can extract the corresponding spatial feature $\F_{n}^{(\ell)} = G_1^{(\ell)}(\z_n)$ for further composition.

%%%% GAN Inversion with Multiple Latent Codes: Adaptive Channel Importance
\vspace{2pt}\noindent\textbf{Adaptive Channel Importance.}
Recall that we would like each $\z_n$ to recover some particular regions of the target image.
Bau \emph{et al.} \cite{gandissection} observed that different units (\emph{i.e.}, channels) of the generator in GAN are responsible for generating different visual concepts such as objects and textures.
Based on this observation, we introduce the adaptive channel importance ${\bm\alpha}_n$ for each $\z_n$ to help them align with different semantics.
Here, ${\bm\alpha}_n\in\R^C$ is a $C$-dimensional vector and $C$ is the number of channels in the $\ell$-th layer of $G(\cdot)$.
We expect each entry of ${\bm\alpha}_n$ to represent how important the corresponding channel of the feature map $\F_{n}^{(\ell)}$ is.
With such composition, the reconstructed image can be generated with
\begin{align}
  \x^{inv} = G_2^{(\ell)}(\sum_{n=1}^N \F_n^{(\ell)} \odot {\bm\alpha}_{n}), \label{eq:feature-composition}
\end{align}
where $\odot$ denotes the channel-wise multiplication as
\begin{align}
  \{\F_n^{(\ell)} \odot {\bm\alpha}_{n}\}_{i,j,c} = \{\F_n^{(\ell)}\}_{i,j,c} \times \{{\bm\alpha}_{n}\}_{c}. \label{eq:adaptive-channel-importance}
\end{align}
Here, $i$ and $j$ indicate the spatial location, while $c$ stands for the channel index.

%%%% GAN Inversion with Multiple Latent Codes: Optimization Objective
\vspace{2pt}\noindent\textbf{Optimization Objective.}
After introducing the feature composition technique together with the introduced adaptive channel importance to integrate multiple latent codes, there are $2N$ sets of parameters to be optimized in total.
Accordingly we reformulate Eq.\eqref{eq:single-code-inversion} as
\begin{align}
  \{\z_n^*\}_{n=1}^N, \{{\bm\alpha}_n^*\}_{n=1}^N = \mathop{\arg\min}_{\{\z_n\}_{n=1}^N,\{{\bm\alpha}_n\}_{n=1}^N} \Loss(\x^{inv}, \x). \label{eq:multi-code-inversion}
\end{align}
To improve the reconstruction quality, we define the objective function by leveraging both low-level and high-level information.
In particular, we use pixel-wise reconstruction error as well as the $l_1$ distance between the perceptual features \cite{johnson2016perceptual} extracted from the two images\footnote{In this experiment, we use pre-trained VGG-16 model \cite{vgg} as the feature extractor, and the output of layer $\rm conv\_43$ is used.}.
Therefore, the objective function is as follows:
\begin{align}
  \Loss(\x_1,\x_2) = ||\x_1-\x_2||_2^2 + ||\phi(\x_1), \phi(\x_2)||_1, \label{eq:objection-function}
\end{align}
where $\phi(\cdot)$ denotes the perceptual feature extractor.
We use the gradient descent algorithm to find the optimal latent codes as well as the corresponding channel importance scores.

%%%% Sub-Section: Multi-Code GAN Prior for Image Processing
\subsection{Multi-Code GAN Prior for Image Processing}\label{subsec:gan-prior-for-image-processing}
%%%%
After inversion, we apply the reconstruction result as multi-code GAN prior to a variety of image processing tasks.
Each task requires an image as a reference, which is the input image for processing.
For example, image colorization task deals with grayscale images and image inpainting task restores images with missing holes.
Given an input, we apply the proposed multi-code GAN inversion method to reconstruct it and then post-process the reconstructed image to approximate the input.
When the approximation is close enough to the input, we assume the reconstruction before post-processing is what we want.
Here, to adapt mGANprior to a specific task, we modify Eq.\eqref{eq:objection-function} based on the post-processing function:
\begin{itemize}
  \vspace{-2pt}
  \setlength{\itemsep}{2pt}
  \setlength{\parsep}{0pt}
  \setlength{\parskip}{0pt}

  \item For image colorization task, with a grayscale image $I_{gray}$ as the input, we expect the inversion result to have the same gray channel as $I_{gray}$ with
  \begin{equation}
    \Loss_{color} = \Loss(\mathtt{gray}(\x^{inv}), I_{gray}), \label{eq:colorization}
  \end{equation}
  where $\mathtt{gray}(\cdot)$ stands for the operation to take the gray channel of an image.

  \item For image super-resolution task, with a low-resolution image $I_{LR}$ as the input, we downsample the inversion result to approximate $I_{LR}$ with
  \begin{equation}
    \Loss_{SR} = \Loss(\mathtt{down}(\x^{inv}), I_{LR}), \label{eq:super-resolution}
  \end{equation}
  where $\mathtt{down}(\cdot)$ stands for the downsampling operation.

  \item For image inpainting task, with an intact image $I_{ori}$ and a binary mask $\mathbf{m}$ indicating known pixels, we only reconstruct the incorrupt parts and let the GAN model fill in the missing pixels automatically with
  \begin{equation}
    \Loss_{inp} = \Loss(\x^{inv} \circ \mathbf{m}, I_{ori} \circ \mathbf{m}), \label{eq:inpainting}
  \end{equation}
  where $\circ$ denotes the element-wise product.

\end{itemize}

%%%% Figure: Inversion Method Comparison
\begin{figure}[t]
  \centering
  \includegraphics[width=1.0\linewidth]{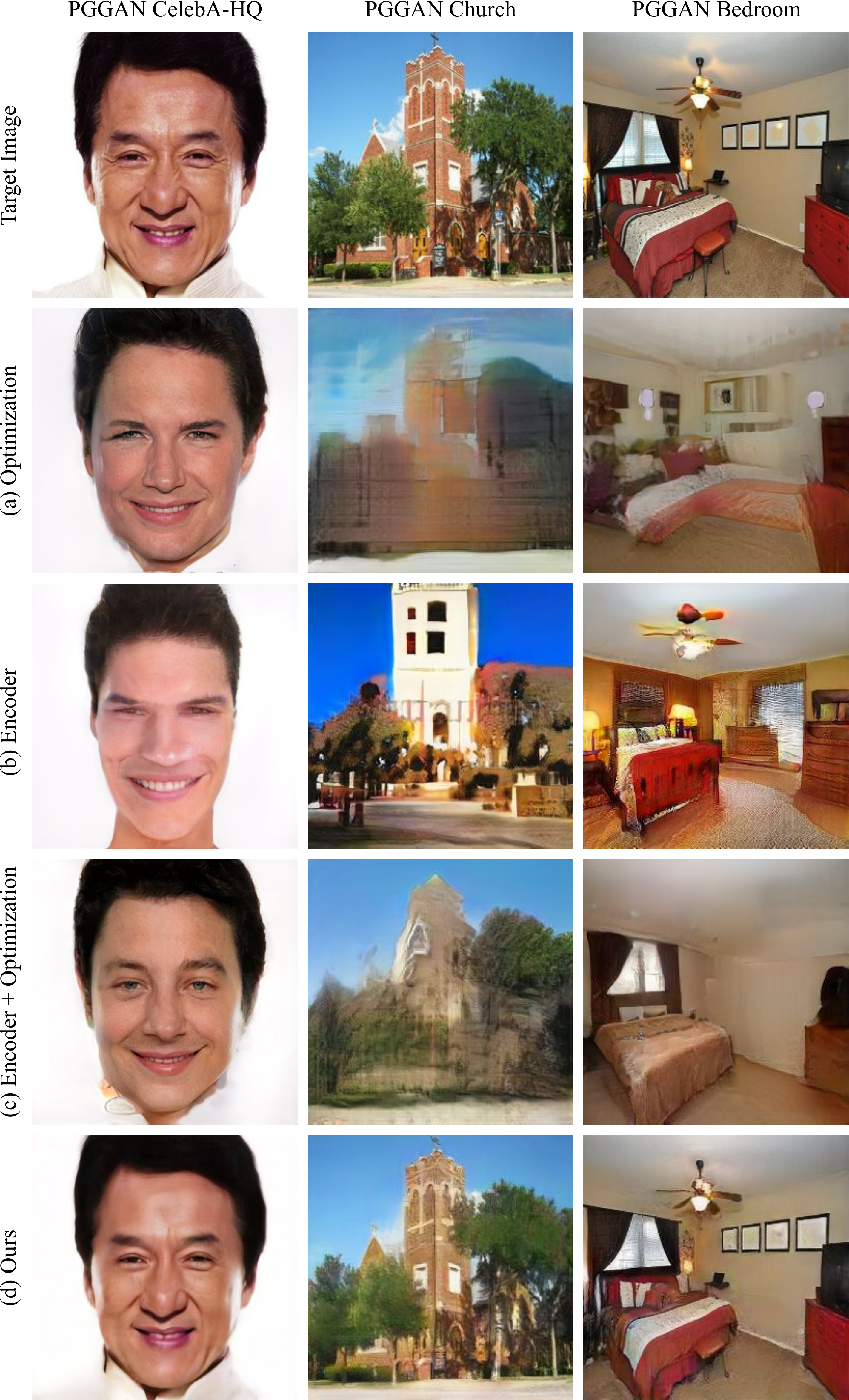}
  \vspace{-20pt}
  \captionsetup{font=small}
  \caption{
    Qualitative comparison of different GAN inversion methods, including (a) optimizing a single latent code \cite{invertibility}, (b) learning an encoder \cite{zhu2016generative}, (c) using the encoder as initialization for optimization \cite{inverting2019}, and (d) our proposed mGANprior.
  }
  \label{fig:inversion}
 \vspace{-12pt}
\end{figure}

%%%% Section: Experiments
\vspace{-6pt}\section{Experiments}\label{sec:experiments}
%%%%
We conduct extensive experiments on state-of-the-art GAN models, \emph{i.e.}, PGGAN \cite{pggan} and StyleGAN \cite{stylegan}, to verify the effectiveness of mGANprior.
These models are trained on various datasets, including CelebA-HQ \cite{pggan} and FFHQ \cite{stylegan} for faces as well as LSUN \cite{lsun} for scenes.

%%%% Table: Inversion Method Comparison
\setlength{\tabcolsep}{4pt}
\begin{table}[t]
  \centering
  \footnotesize
  \captionsetup{font=small}
  \caption{
    Quantitative comparison of different GAN inversion methods: including (a) optimizing a single latent code \cite{invertibility}, (b) learning an encoder \cite{zhu2016generative}, (c) using the encoder as initialization for optimization \cite{inverting2019}, and (d) our proposed mGANprior.
    $\uparrow$ means the higher the better while $\downarrow$ means the lower the better.
  }
  \label{tab:inversion}
  \vspace{-8pt}
  \begin{tabular}{ccccccc}
    \hline
            & \multicolumn{2}{c}{Bedroom} & \multicolumn{2}{c}{Church} & \multicolumn{2}{c}{Face} \\
    \cmidrule(lr){2-3} \cmidrule(lr){4-5} \cmidrule(lr){6-7}
    Method  & PSNR$\uparrow$ & LPIPS$\downarrow$ & PSNR$\uparrow$ & LPIPS$\downarrow$ & PSNR$\uparrow$ & LPIPS$\downarrow$ \\
    \hline
    (a)
    & 17.19 & 0.5897 & 17.15 & 0.5339 & 19.17 & 0.5797 \\
    (b)
    & 11.59 & 0.6247 & 11.58 & 0.5961 &  11.18 & 0.6992 \\
    (c)
    & 18.34 & 0.5201 & 17.81 & 0.4789 &  20.33 & 0.5321 \\
    (d)
    & \bf 25.13 & \bf 0.1578 & \bf 22.76 & \bf 0.1799 & \bf 23.59 & \bf 0.4432 \\
    \hline
  \end{tabular}
  \vspace{-10pt}
\end{table}

%%%% Sub-Section: Comparison with Other Inversion Methods
\subsection{Comparison with Other Inversion Methods}\label{subsec:inversion-comparison}
%%%%
There are many attempts on GAN inversion in the literature.
In this section, we compare our multi-code inversion approach with the following baseline methods:
(a) optimizing a single latent code $\z$ as in Eq.\eqref{eq:single-code-inversion} \cite{invertibility},
(b) learning an encoder to reverse the generator \cite{zhu2016generative},
and (c) combing (a) and (b) by using the output of the encoder as the initialization for further optimization \cite{inverting2019}.

To quantitatively evaluate the inversion results, we introduce the Peak Signal-to-Noise Ratio (PSNR) to measure the similarity between the original input and the reconstruction result from pixel level, as well as the LPIPS metric \cite{zhang2018unreasonable} which is known to align with human perception.
We make comparisons on three PGGAN \cite{pggan} models that are trained on LSUN bedroom (indoor scene), LSUN church (outdoor scene), and CelebA-HQ (human face) respectively.
For each model, we invert 300 real images for testing.

Tab.\ref{tab:inversion} and Fig.\ref{fig:inversion} show the quantitative and qualitative comparisons respectively.
From Tab.\ref{tab:inversion}, we can tell that mGANprior beats other competitors on all three models from both pixel level (PSNR) and perception level (LPIPS).
We also observe in Fig.\ref{fig:inversion} that existing methods fail to recover the details of the target image, which is due to the limited representation capability of a single latent code.
By contrast, our method achieves much more satisfying reconstructions with most details, benefiting from multiple latent codes.
We even recover an eastern face with a model trained on western data (CelebA-HQ \cite{pggan}).

%%%% Sub-Section: Analysis on Inverted Codes
\subsection{Analysis on Inverted Codes}\label{subsec:analysis-on-inverted-codes}
As described in Sec.\ref{sec:method}, our method achieves high-fidelity GAN inversion with $N$ latent codes and $N$ importance factors.
Taking PGGAN as an example, if we choose the 6th layer (\emph{i.e.}, with 512 channels) as the composition layer with $N=10$, the number of parameters to optimize is $10\times(512+512)$, which is 20 times the dimension of the original latent space.
In this section, we perform detailed analysis on the inverted codes.

\begin{figure}[t]
  \centering
  \includegraphics[width=1.0\linewidth]{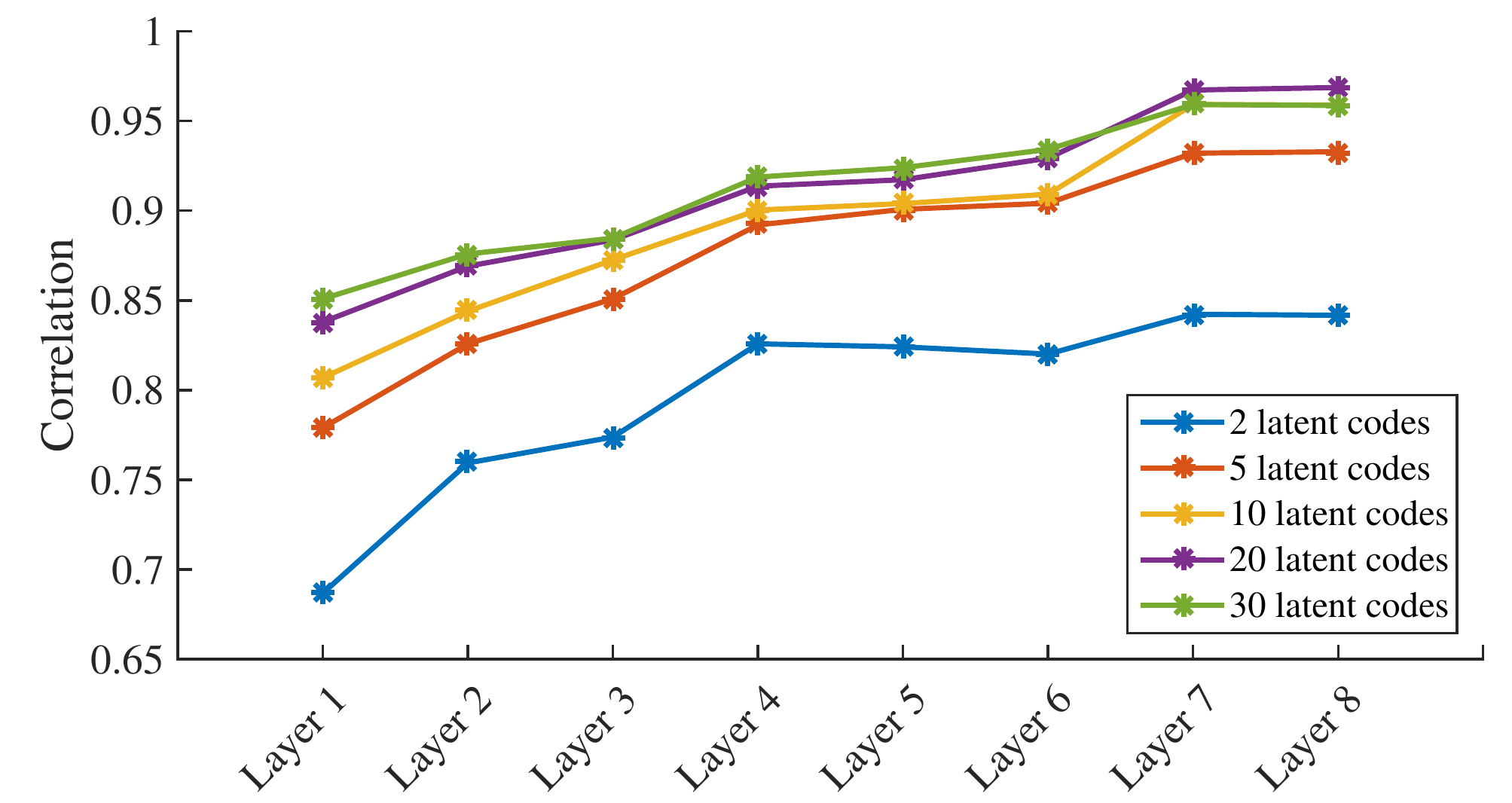}
  \vspace{-20pt}
  \captionsetup{font=small}
  \caption{
    Effects on inversion performance by the number of latent codes used and the feature composition position.
  }
  \label{fig:ablation}
  \vspace{-8pt}
\end{figure}

\vspace{2pt}\noindent\textbf{Number of Codes.}
Obviously, there is a trade-off between the dimension of the optimization space and the inversion quality.
To better analysis such trade-off, we evaluate our method by varying the number of latent codes to optimize.
Fig.\ref{fig:ablation} shows that the more latent codes used, the better reconstruction we are able to obtain.
However, it does not imply that the performance can be infinitely improved by increasing the number of latent codes.
From Fig.\ref{fig:ablation}, we can see that after the number reaches 20, there is no significant improvement via involving more latent codes.

%%%% Figure: Ablation

%%%% Figure: Role of z.
\begin{figure}[t]
  \centering
  \includegraphics[width=1.0\linewidth]{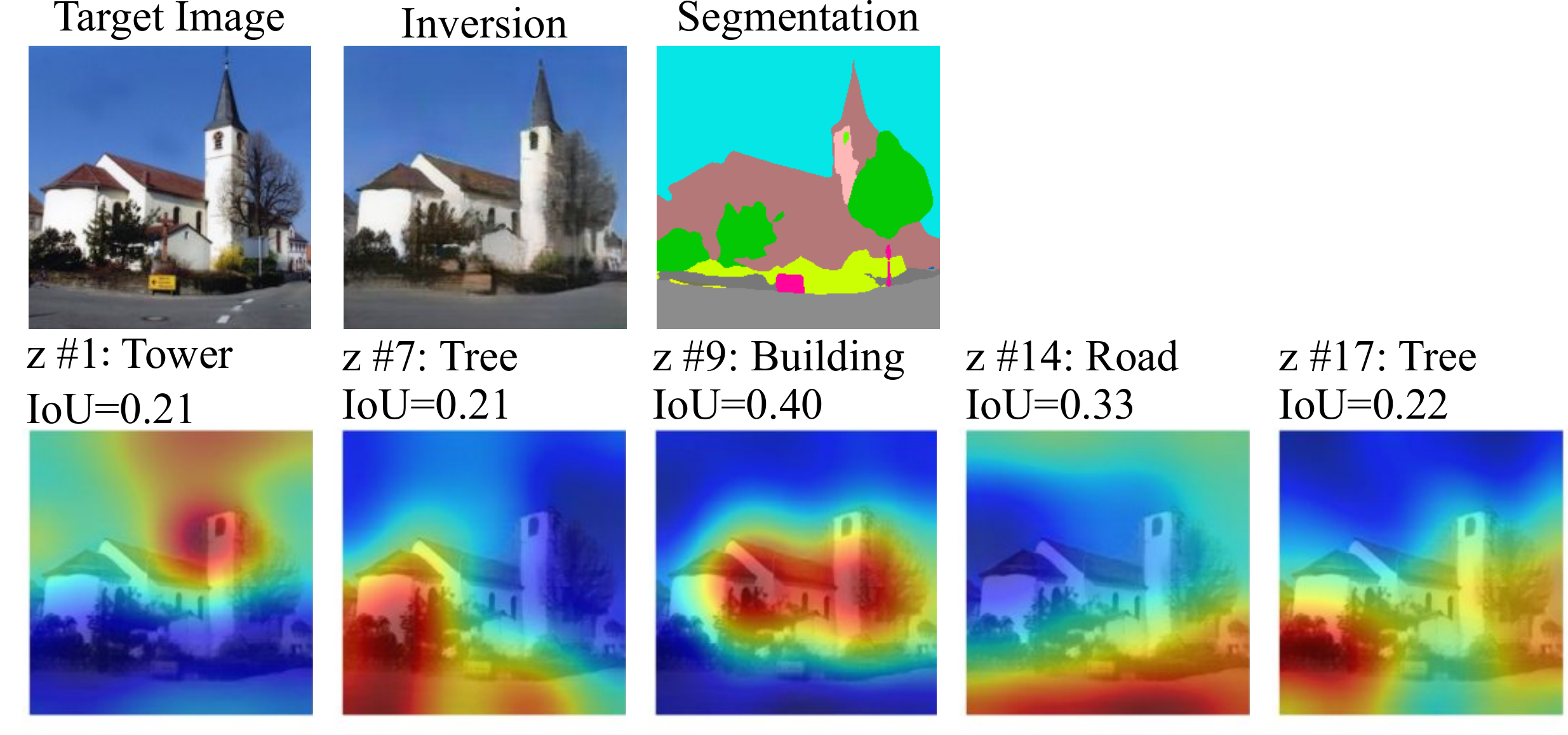}
  \vspace{-20pt}
  \captionsetup{font=small}
  \caption{
    Visualization of the role of each latent code.
    On the top row are the target image, inversion result, and the corresponding segmentation mask, respectively.
    On the bottom row are several latent codes annotated with a specific semantic label.
  }
  \label{fig:role-of-z}
  \vspace{-15pt}
\end{figure}

\vspace{2pt}\noindent\textbf{Different Composition Layers.}
On which layer to perform feature composition also affects the performance of the proposed mGANprior.
We thus compose the latent codes on various layers of PGGAN (\emph{i.e.}, from 1st to 8th) and compare the inversion quality, as shown in Fig.\ref{fig:ablation}.
In general, a higher composition layer could lead to a better inversion effect.
However, as revealed in \cite{gandissection}, higher layers contain the information of local pixel patterns such as edges and colors rather than the high-level semantics.
Composing features at higher layers is hard to reuse of the semantic knowledge learned by GANs.
This will be discussed more in Sec.\ref{subsec:knowledge-representation-in-gans}.

%%%% Figure: Colorization
\begin{figure*}[t]
  \centering
  \includegraphics[width=1.0\linewidth]{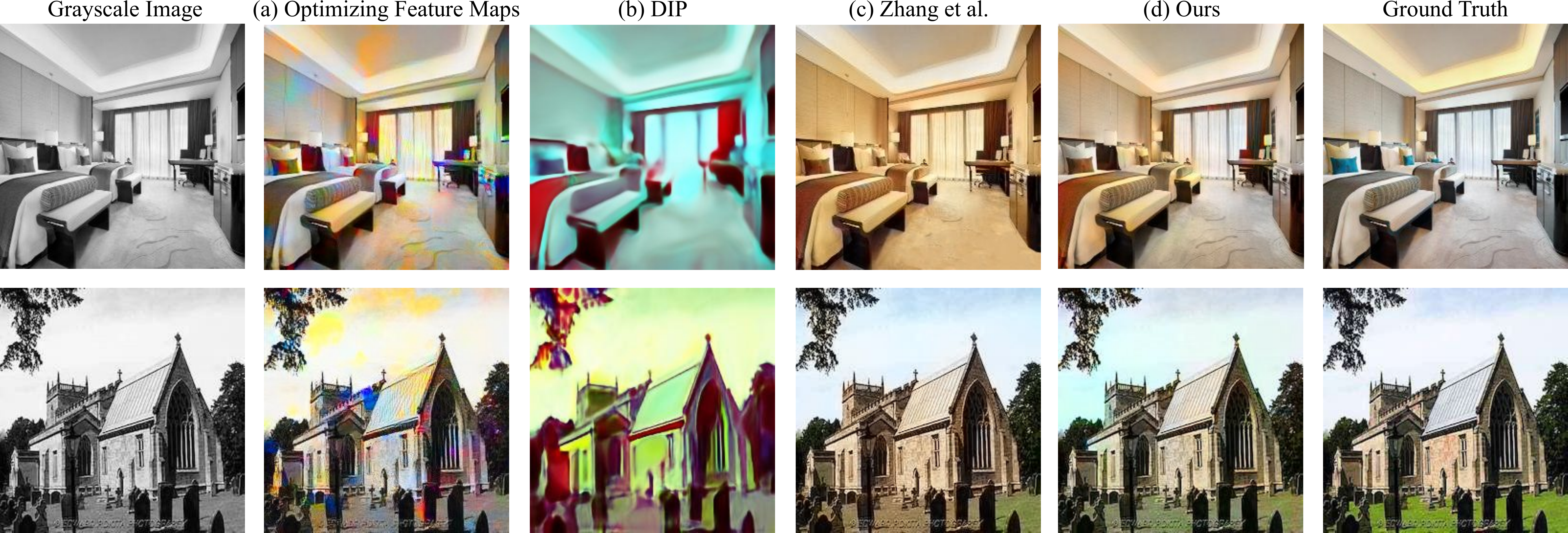}
  \vspace{-20pt}
  \captionsetup{font=small}
  \caption{
    Qualitative comparison of different colorization methods, including (a) inversion by optimizing feature maps \cite{bau2019semantic}, (b) DIP \cite{ulyanov2018deep}, (c) Zhang \emph{et al.} \cite{zhang2016colorful}, and (d) our mGANprior.
  }
  \label{fig:colorization}
 \vspace{-15pt}
\end{figure*}

\vspace{2pt}\noindent\textbf{Role of Each Latent Code.}
We employ multiple latent codes by expecting each of them to take charge of inverting a particular region and hence complement with each other.
In this part, we visualize the roles that different latent codes play in the inversion process.
As pointed out by \cite{gandissection}, for a particular layer in a GAN model, different units (channels) control different semantic concepts.
Recall that mGANprior uses adaptive channel importance to help determine what kind of semantics a particular $\z$ should focus on.
Therefore, for each $\z_n$, we set the elements in $\bm\alpha_n$ that are larger than 0.2 as 0, getting $\bm\alpha'_n$.
Then we compute the difference map between the reconstructions using $\bm\alpha_n$ and $\bm\alpha'_n$.
With the help of a segmentation model \cite{zhou2017scene}, we can also get the segmentation maps for various visual concepts, such as tower and tree.
We finally annotate each latent code based on the Intersection-over-Union (IoU) metric between the corresponding difference map and all candidate segmentation maps.
Fig.\ref{fig:role-of-z} shows the segmentation result and the IoU maps of some chosen latent codes.
It turns out that the latent codes are specialized to invert different meaningful image regions to compose the whole image.
This is also a huge advantage of using multiple latent codes over using a single code.

%%%% Table: Colorization
\setlength{\tabcolsep}{10pt}
\begin{table}[t]
  \centering
  \footnotesize
  \captionsetup{font=small}
  \caption{
    Quantitative evaluation results on colorization task with bedroom and church images.
    AuC refers to the area under the curve of the cumulative error distribution over \emph{ab} color space \cite{zhang2016colorful}.
    $\uparrow$ means higher score is better.
  }
  \label{tab:colorization}
  \vspace{-10pt}
  \begin{tabular}{lccc}
    \hline
                                                       & Bedroom            & Church \\
    Method                                             & AuC (\%)$\uparrow$ & AuC (\%)$\uparrow$ \\
    \hline
    Grayscale input                                    & 88.02              & 85.50 \\
    % (a) Optimizing a single latent code              & 85.87              & 85.93 \\
    (a) Optimizing feature maps \cite{bau2019semantic} & 85.41              & 86.10 \\
    (b) DIP \cite{ulyanov2018deep}                     & 84.33              & 83.31 \\
    (c) Zhang \emph{et al.} \cite{zhang2016colorful}   & 88.55              & 89.13 \\
    (d) Ours                                           & \bf 90.02          & \bf 89.43 \\
    \hline
  \end{tabular}
  \vspace{-10pt}
\end{table}

%%%% Sub-Section: Image Processing Applications
\subsection{Image Processing Applications}\label{subsec:application}
%%%%
With the high-fidelity image reconstruction, our multi-code inversion method facilitates many image processing tasks with \emph{pre-trained} GANs as prior.
In this section, we apply the proposed mGANprior to a variety of real-world applications to demonstrate its effectiveness, including image colorization, image super-resolution, image inpainting and denoising, as well as semantic manipulation and style mixing.
For each application, the GAN model is \emph{fixed}.

%%%% Image Processing Applications: Image Colorization
\vspace{2pt}\noindent\textbf{Image Colorization.}
Given a grayscale image as input, we can colorize it with mGANprior as described in Sec.\ref{subsec:gan-prior-for-image-processing}.
We compare our inversion method with optimizing the intermediate feature maps \cite{bau2019semantic}.
We also compare with DIP \cite{ulyanov2018deep}, which uses a discriminative model as prior, and Zhang \emph{et al.} \cite{zhang2016colorful}, which is specially designed for colorization task.
We do experiments on PGGAN models trained for bedroom and church synthesis, and use the area under the curve of the cumulative error distribution over \emph{ab} color space as the evaluation metric, following \cite{zhang2016colorful}.
Tab.\ref{tab:colorization} and Fig.\ref{fig:colorization} show the quantitative and qualitative comparisons respectively.
It turns out that using the discriminative model as prior fails to colorize the image adequately.
That is because discriminative models focus on learning high-level representation which are not suitable for low-level tasks.
On the contrary, using the generative model as prior leads to much more satisfying colorful images.
We also achieve comparable results as the model whose primary goal is image colorization (Fig.\ref{fig:colorization} (c) and (d)).
This benefits from the rich knowledge learned by GANs.
Note that Zhang \emph{et al.} \cite{zhang2016colorful} is proposed for general image colorization, while our approach can be only applied to a certain image category corresponding to the given GAN model.
A larger GAN model trained on a more diverse dataset should improve its generalization ability.

%%%% Figure: Super-Resolution
\begin{figure}[t]
  \centering
  \includegraphics[width=1.0\linewidth]{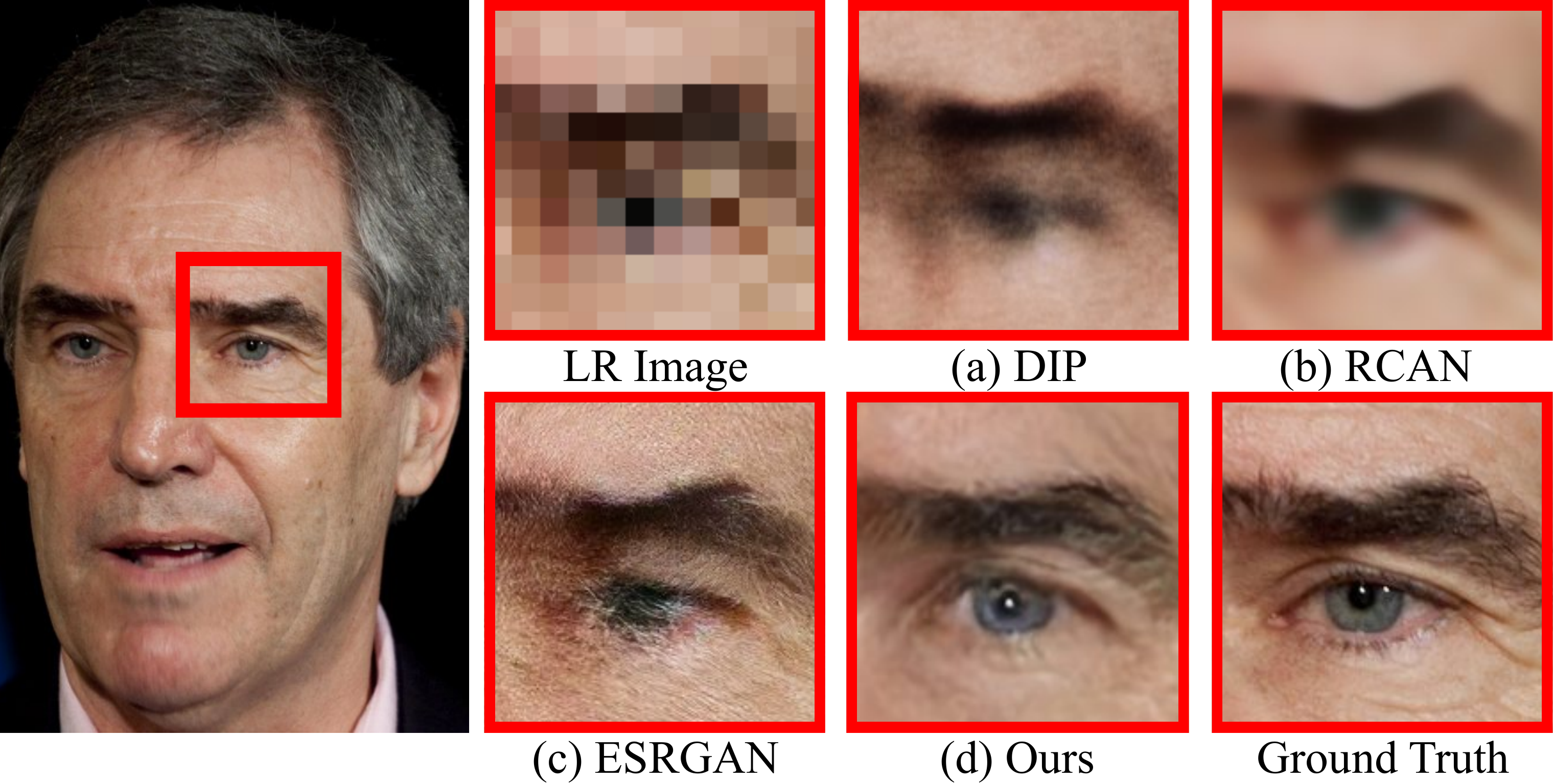}
  \vspace{-20pt}
  \captionsetup{font=small}
  \caption{
    Qualitative comparison of different super-resolution methods with SR factor 16.
    Competitors include DIP \cite{ulyanov2018deep}, RCAN \cite{rcan}, and ESRGAN \cite{wang2018esrgan}.
  }
  \label{fig:super-resolution}
  \vspace{-15pt}
\end{figure}

%%%% Figure: Inpainting
\begin{figure*}[t]
  \centering
  \includegraphics[width=0.98\linewidth]{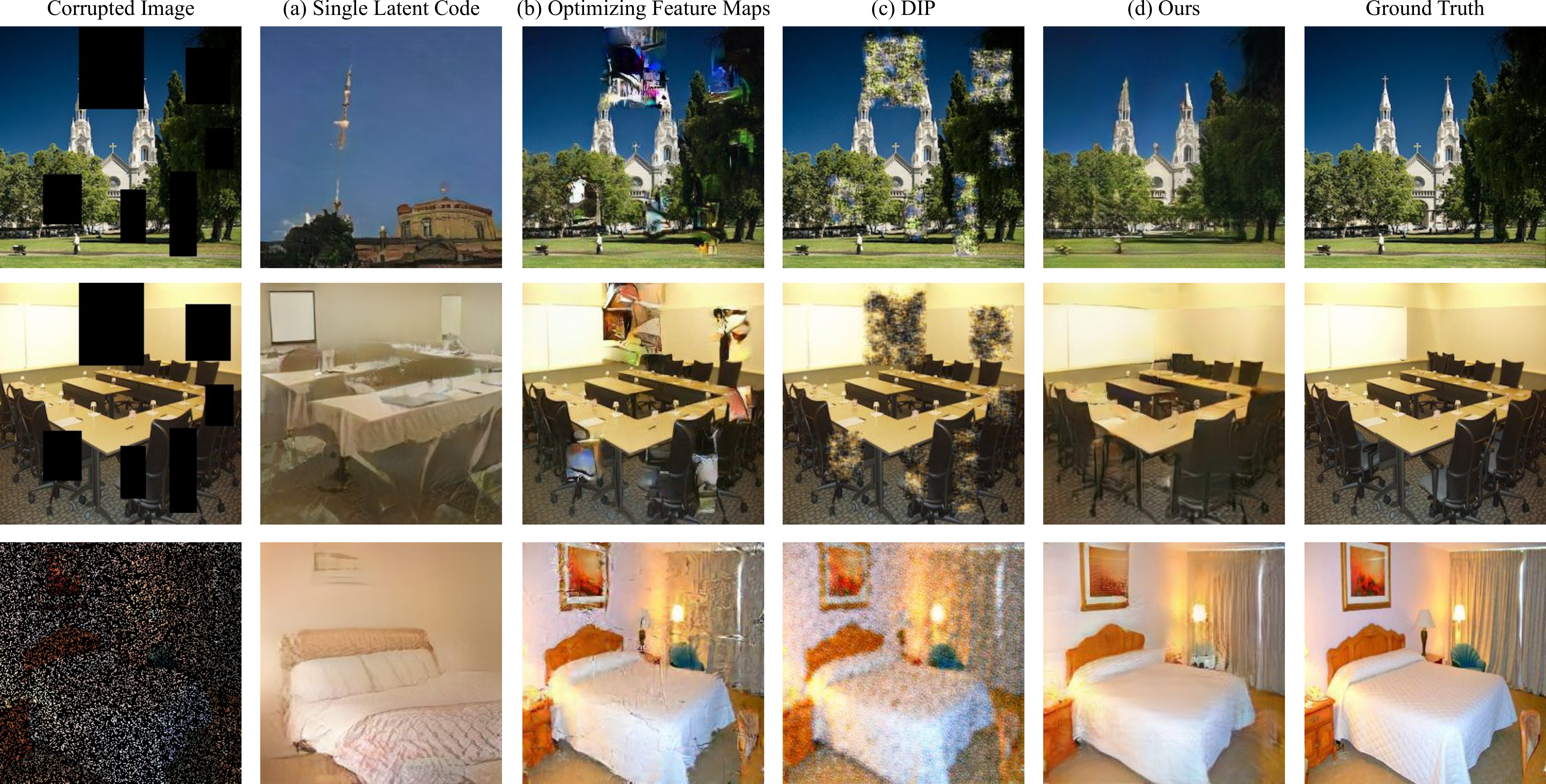}
  \vspace{-8pt}
  \captionsetup{font=small}
  \caption{
    Qualitative comparison of different inpainting methods, including (a) inversion by optimizing a single latent code \cite{lipton2017precise,invertibility}, (b) inversion by optimizing feature maps \cite{bau2019semantic}, (c) DIP \cite{ulyanov2018deep}, and (d) our mGANprior.
  }
  \label{fig:inpainting}
 \vspace{-7pt}
\end{figure*}

%%%% Figure: Manipulation
\begin{figure*}[t]
  \centering
  \includegraphics[width=0.98\linewidth]{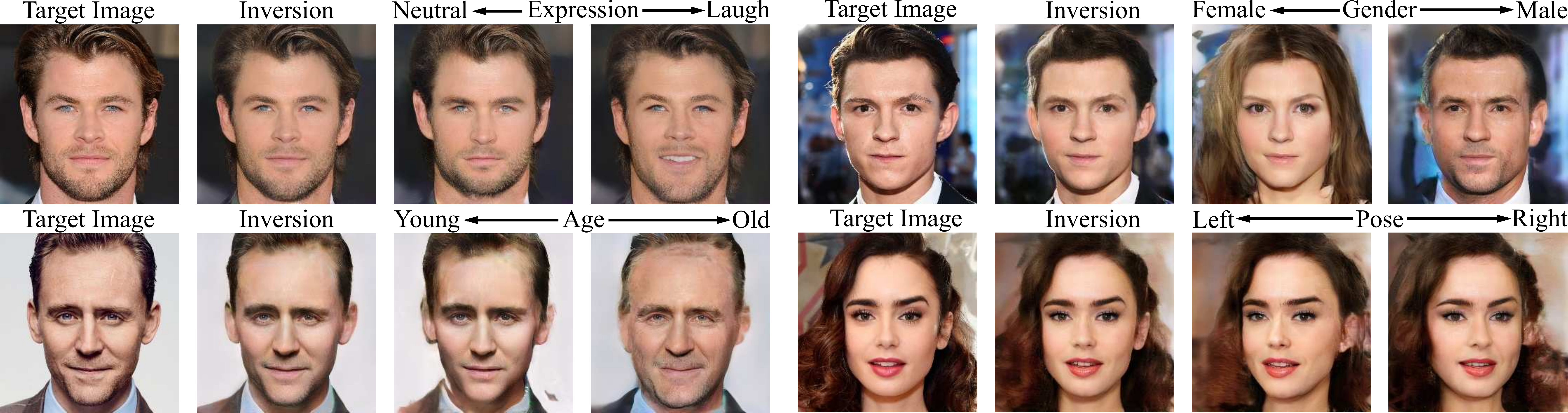}
  \vspace{-8pt}
  \captionsetup{font=small}
  \caption{
    Real face manipulation with respect to four various attributes. In each four-element tuple, from left to right are: input face, inversion result, and manipulation results by making a particular semantic more negative and more positive.
  }
  \label{fig:manipulation}
 \vspace{-15pt}
\end{figure*}

%%%% Image Processing Applications: Image Super-Resolution
\vspace{2pt}\noindent\textbf{Image Super-Resolution.}
We also evaluate our approach on the image super-resolution (SR) task.
We do experiments on the PGGAN model trained for face synthesis and set the SR factor as 16.
Such a large factor is very challenging for the SR task.
We compare with DIP \cite{ulyanov2018deep} as well as the state-of-the-art SR methods, RCAN \cite{rcan} and ESRGAN \cite{wang2018esrgan}.
Besides PSNR and LPIPS, we introduce Naturalness Image Quality Evaluator (NIQE) \cite{niqe} as an extra metric.
Tab.\ref{tab:super-resolution} shows the quantitative comparison.
We can conclude that our approach achieves comparable or even better performance than the advanced learning-based competitors.
A visualization example is also shown in Fig.\ref{fig:super-resolution}, where our method reconstructs the human eye with more details.
Compared to existing learning-based models, like RCAN and ESRGAN, our mGANprior is more flexible to the SR factor.
This suggests that the freely-trained PGGAN model has spontaneously learned rich knowledge such that it can be used as prior to enhance a low-resolution (LR) image.

%%%% Table: Super-Resolution
\setlength{\tabcolsep}{12pt}
\begin{table}[t]
  \centering
  \footnotesize
  \captionsetup{font=small}
  \caption{
    Quantitative comparison of different super-resolution methods with SR factor 16.
    Competitors include DIP \cite{ulyanov2018deep}, RCAN \cite{rcan}, and ESRGAN \cite{wang2018esrgan}.
    $\uparrow$ means the higher the better while $\downarrow$ means the lower the better.
  }
  \label{tab:super-resolution}
  \vspace{-8pt}
  \begin{tabular}{lccc}
    \hline
    Method                           & PSNR$\uparrow$ & LPIPS$\downarrow$ & NIQE$\downarrow$ \\
    \hline
    (a) DIP \cite{ulyanov2018deep}   & 26.87          & 0.4236            & 4.66 \\
    (b) RCAN \cite{rcan}             & \bf 28.82      & 0.4579            & 5.70 \\
    (c) ESRGAN \cite{wang2018esrgan} & 25.26          & 0.3862            & 3.27 \\
    (d) Ours                         & 26.93          & \bf 0.3584        & \bf 3.19 \\
    \hline
  \end{tabular}
  \vspace{-5pt}
\end{table}

%%%% Table: Inpainting
\setlength{\tabcolsep}{3.5pt}
\begin{table}[t]
  \centering
  \footnotesize
  \captionsetup{font=small}
  \caption{
    Quantitative comparison of different inpainting methods.
    We do test with both centrally cropping a $64\times64$ box and randomly cropping 80\% pixels.
    $\uparrow$ means higher score is better.
  }
  \label{tab:inpainting}
  \vspace{-8pt}
  \begin{tabular}{lcccc}
    \hline
    & \multicolumn{2}{c}{Center Crop}  & \multicolumn{2}{c}{Random Crop} \\
    \cmidrule(lr){2-3} \cmidrule(lr){4-5}
    Method                                                        & PSNR$\uparrow$ & SSIM$\uparrow$ & PSNR$\uparrow$ & SSIM$\uparrow$ \\
    \hline
    (a) Single latent code \cite{lipton2017precise,invertibility} & 10.37           & 0.1672         & 12.79          & 0.1783 \\
    (b) Optimizing feature maps \cite{bau2019semantic}            & 14.75           & 0.4563         & 18.72          & 0.2793 \\
    (c) DIP \cite{ulyanov2018deep}                                & 17.92           & 0.4327         & 18.02          & 0.2823 \\
    (d) Ours                                                      & \bf 21.43       & \bf 0.5320     & \bf 22.11      & \bf 0.5532 \\
    \hline
  \end{tabular}
  \vspace{-25pt}
\end{table}

%%%% Image Processing Applications: Image Inpainting and Denoising
\vspace{2pt}\noindent\textbf{Image Inpainting and Denoising.}
We further extend our approach to image restoration tasks, like image inpainting and image denoising.
We first corrupt the image contents by randomly cropping or adding noises, and then use different algorithms to restore them.
Experiments are conducted on PGGAN models and we compare with several baseline inversion methods as well as DIP \cite{ulyanov2018deep}.
PSNR and Structural SIMilarity (SSIM) \cite{ssim} are used as evaluation metrics.
Tab.\ref{tab:inpainting} shows the quantitative comparison, where our approach achieves the best performances on both settings of center crop and random crop.
Fig.\ref{fig:inpainting} includes some examples of restoring corrupted images.
It is obvious that both existing inversion methods and DIP fail to adequately fill in the missing pixels or completely remove the added noises.
By contrast, our method is able to use well-trained GANs as prior to convincingly repair the corrupted images with meaningful filled content.

%%%% Image Processing Applications: Semantic Manipulation
\vspace{2pt}\noindent\textbf{Semantic Manipulation.}
Besides the aforementioned low-level applications, we also test our approach with some high-level tasks, like semantic manipulation and style mixing.
As pointed out by prior work \cite{jahanian2020steerability,goetschalckx2019ganalyze,shen2020interpreting}, GANs have already encoded some interpretable semantics inside the latent space.
From this point, our inversion method provides a feasible way to utilize these learned semantics for \emph{real} image manipulation.
We apply the manipulation framework based on latent code proposed in \cite{shen2020interpreting} to achieve semantic facial attribute editing.
Fig.\ref{fig:manipulation} shows the manipulation results.
We see that mGANprior can provide rich enough information for semantic manipulation.

%%%% Figure: Different Prior
\begin{figure}[t]
  \centering
  \includegraphics[width=1.0\linewidth]{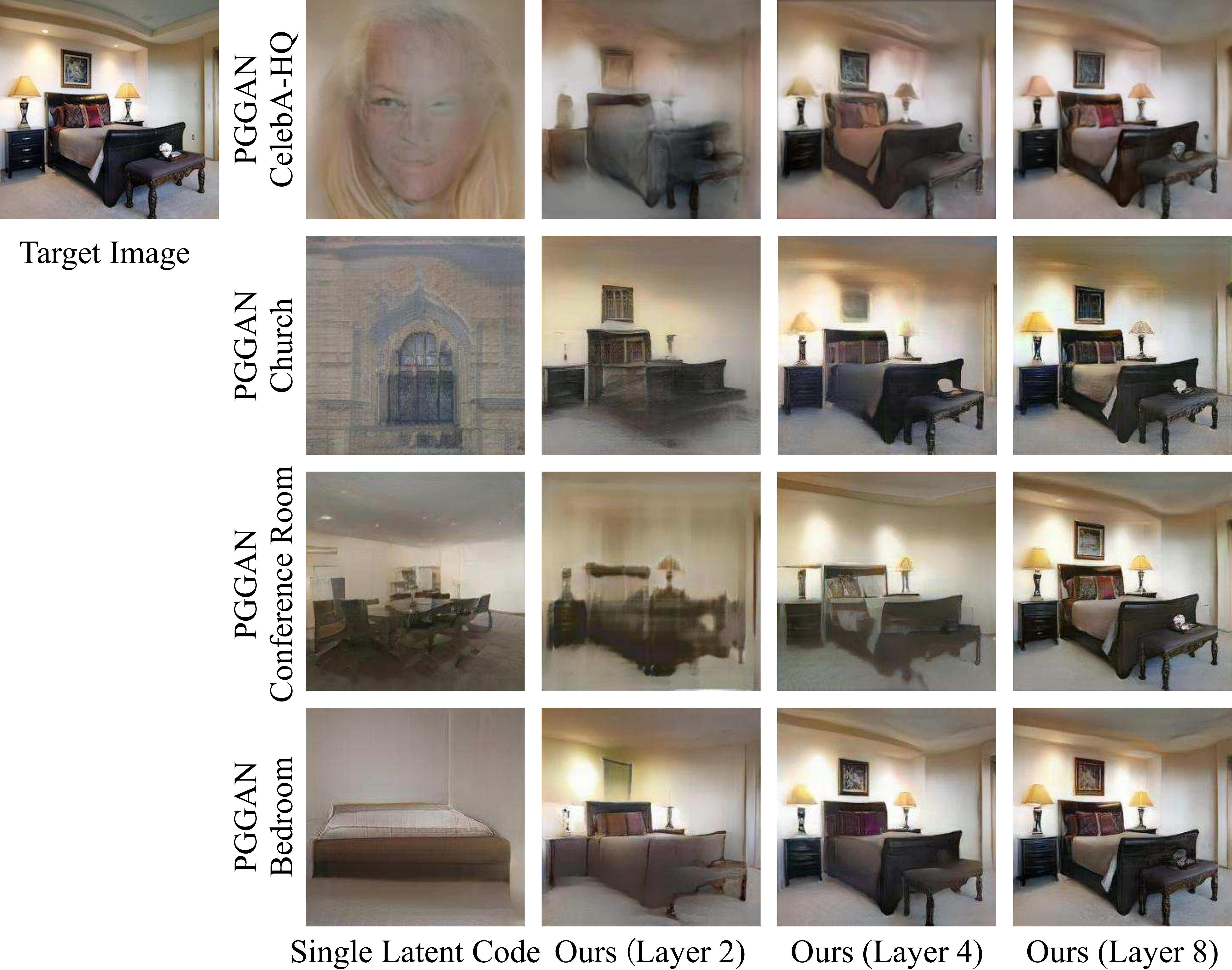}
  \vspace{-20pt}
  \captionsetup{font=small}
  \caption{
    Comparison of the inversion results using different GAN models as well as performing feature composition at different layers.
    Each row stands for a PGGAN model trained on a specific dataset as prior, while each column shows results by composing feature maps at a certain layer.
  }
  \label{fig:different-prior}
  \vspace{-15pt}
\end{figure}

%%%% Figure: Layer-wise Analysis
\begin{figure}[t]
  \centering
  \includegraphics[width=1.0\linewidth]{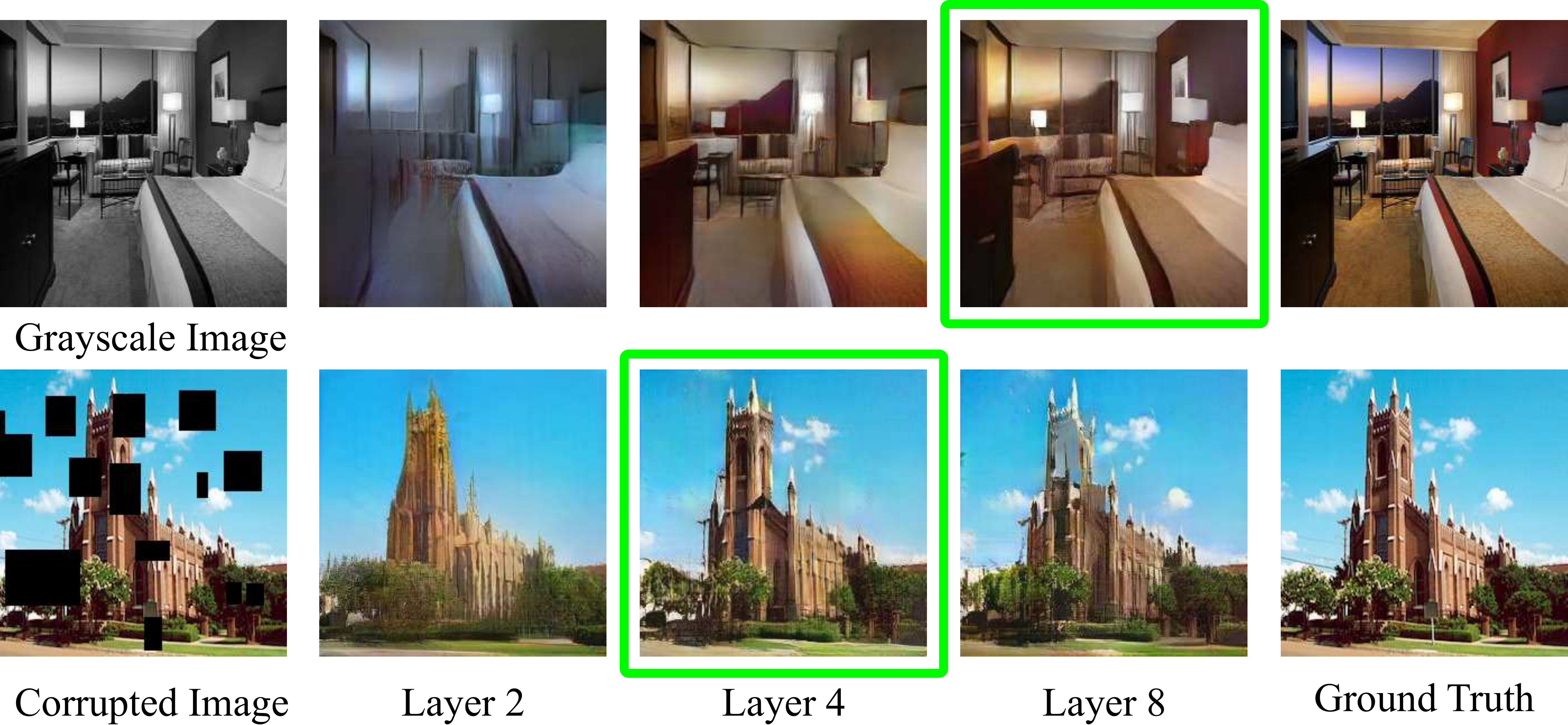}
  \vspace{-20pt}
  \captionsetup{font=small}
  \caption{
    Colorization and inpainting results with mGANprior using different composition layers.
    AuC (the higher the better) for colorization task are 86.83\%, 87.44\%, 90.02\% with respect to the 2nd, 4th, and 8th layer respectively.
    PSNR (the higher the better) for inpainting task are 21.19db, 22.11db, 20.70db with respect to the 2nd, 4th, and 8th layer respectively.
    Images in \textcolor{green}{\textbf{green}} boxes indicate the best results.
  }
  \label{fig:layer-wise-analysis}
 \vspace{-16pt}
\end{figure}

%%%% Sub-Section: Knowledge Representation in GANs
\subsection{Knowledge Representation in GANs}\label{subsec:knowledge-representation-in-gans}
As discussed above, the major limitation of using single latent code is its limited expressiveness, especially when the test image presents domain gap to the training data.
Here we verify whether using multiple codes can help alleviate this problem.
In particular, we try to use GAN models trained for synthesizing face, church, conference room, and bedroom, to invert a bedroom image.
As shown in Fig.\ref{fig:different-prior}, when using a single latent code, the reconstructed image still lies in the original training domain (\emph{e.g.}, the inversion with PGGAN CelebA-HQ model looks like a face instead of a bedroom).
On the contrary, our approach is able to compose a bedroom image no matter what data the GAN generator is trained with.

We further analyze the layer-wise knowledge of a well-trained GAN model by performing feature composition at different layers.
Fig.\ref{fig:different-prior} suggests that the higher layer is used, the better the reconstruction will be.
That is because reconstruction focuses on recovering low-level pixel values, and GANs tend to represent abstract semantics at bottom layers while represent content details at top layers.
We also observe that the 4th layer is good enough for the bedroom model to invert a bedroom image, but the other three models need the 8th layer for satisfying inversion.
The reason is that bedroom shares different semantics from face, church, and conference room, therefore the high-level knowledge (contained in bottom layers) from these models cannot be reused.
We further make per-layer analysis by applying our approach to image colorization and image inpainting tasks, as shown in Fig.\ref{fig:layer-wise-analysis}.
The colorization task gets the best result at the 8th layer while the inpainting task at the 4th layer.
That is because colorization is more like a low-level rendering task while inpainting requires the GAN prior to fill in the missing content with meaningful objects.
This is consistent with the analysis from Fig.\ref{fig:different-prior}, which is that low-level knowledge from GAN prior can be reused at higher layers while high-level knowledge at lower layers.

%%%% Section: Conclusion
\section{Conclusion}\label{sec:conclusion}
%%%%
We present mGANprior that employs multiple latent codes for reconstructing real images with a pre-trained GAN model.
It enables these GAN models as powerful prior to a variety of image processing tasks.

\vspace{1pt}\noindent\textbf{Acknowledgement:}
This work is supported in part by the Early Career Scheme (ECS) through the Research Grants Council of Hong Kong under Grant No.24206219 and in part by SenseTime Collaborative Grant.

{\small
\bibliographystyle{ieee_fullname}
\bibliography{ref}

\begin{thebibliography}{10}\itemsep=-1pt

\bibitem{wgan}
Martin Arjovsky, Soumith Chintala, and L{\'e}on Bottou.
\newblock Wasserstein gan.
\newblock {\em arXiv preprint arXiv:1701.07875}, 2017.

\bibitem{athar2019latent}
ShahRukh Athar, Evgeny Burnaev, and Victor Lempitsky.
\newblock Latent convolutional models.
\newblock In {\em ICLR}, 2019.

\bibitem{bau2019semantic}
David Bau, Hendrik Strobelt, William Peebles, Jonas Wulff, Bolei Zhou, Jun-Yan
  Zhu, and Antonio Torralba.
\newblock Semantic photo manipulation with a generative image prior.
\newblock In {\em SIGGRAPH}, 2019.

\bibitem{gandissection}
David Bau, Jun-Yan Zhu, Hendrik Strobelt, Bolei Zhou, Joshua~B. Tenenbaum,
  William~T. Freeman, and Antonio Torralba.
\newblock Gan dissection: Visualizing and understanding generative adversarial
  networks.
\newblock In {\em ICLR}, 2019.

\bibitem{inverting2019}
David Bau, Jun-Yan Zhu, Jonas Wulff, William Peebles, Hendrik Strobelt, Bolei
  Zhou, and Antonio Torralba.
\newblock Inverting layers of a large generator.
\newblock In {\em ICLR Workshop}, 2019.

\bibitem{bau2019seeing}
David Bau, Jun-Yan Zhu, Jonas Wulff, William Peebles, Hendrik Strobelt, Bolei
  Zhou, and Antonio Torralba.
\newblock Seeing what a gan cannot generate.
\newblock In {\em ICCV}, 2019.

\bibitem{began}
David Berthelot, Thomas Schumm, and Luke Metz.
\newblock Began: Boundary equilibrium generative adversarial networks.
\newblock {\em arXiv preprint arXiv:1703.10717}, 2017.

\bibitem{biggan}
Andrew Brock, Jeff Donahue, and Karen Simonyan.
\newblock Large scale gan training for high fidelity natural image synthesis.
\newblock In {\em ICLR}, 2019.

\bibitem{chen2018image}
Jingwen Chen, Jiawei Chen, Hongyang Chao, and Ming Yang.
\newblock Image blind denoising with generative adversarial network based noise
  modeling.
\newblock In {\em CVPR}, 2018.

\bibitem{chen2018gated}
Xinyuan Chen, Chang Xu, Xiaokang Yang, Li Song, and Dacheng Tao.
\newblock Gated-gan: Adversarial gated networks for multi-collection style
  transfer.
\newblock {\em TIP}, 2018.

\bibitem{stargan}
Yunjey Choi, Minje Choi, Munyoung Kim, Jung-Woo Ha, Sunghun Kim, and Jaegul
  Choo.
\newblock Stargan: Unified generative adversarial networks for multi-domain
  image-to-image translation.
\newblock In {\em CVPR}, 2018.

\bibitem{inverting2018}
Antonia Creswell and Anil~Anthony Bharath.
\newblock Inverting the generator of a generative adversarial network.
\newblock {\em TNNLS}, 2018.

\bibitem{donahue2016adversarial}
Jeff Donahue, Philipp Kr{\"a}henb{\"u}hl, and Trevor Darrell.
\newblock Adversarial feature learning.
\newblock In {\em ICLR}, 2017.

\bibitem{dumoulin2016adversarially}
Vincent Dumoulin, Ishmael Belghazi, Ben Poole, Olivier Mastropietro, Alex Lamb,
  Martin Arjovsky, and Aaron Courville.
\newblock Adversarially learned inference.
\newblock In {\em ICLR}, 2017.

\bibitem{goetschalckx2019ganalyze}
Lore Goetschalckx, Alex Andonian, Aude Oliva, and Phillip Isola.
\newblock Ganalyze: Toward visual definitions of cognitive image properties.
\newblock In {\em ICCV}, 2019.

\bibitem{gan}
Ian Goodfellow, Jean Pouget-Abadie, Mehdi Mirza, Bing Xu, David Warde-Farley,
  Sherjil Ozair, Aaron Courville, and Yoshua Bengio.
\newblock Generative adversarial nets.
\newblock In {\em NeurIPS}, 2014.

\bibitem{wgangp}
Ishaan Gulrajani, Faruk Ahmed, Martin Arjovsky, Vincent Dumoulin, and Aaron~C
  Courville.
\newblock Improved training of wasserstein gans.
\newblock In {\em NeurIPS}, 2017.

\bibitem{hand2019global}
Paul Hand and Vladislav Voroninski.
\newblock Global guarantees for enforcing deep generative priors by empirical
  risk.
\newblock {\em IEEE Transactions on Information Theory}, 2019.

\bibitem{hao2018mixgan}
Guang-Yuan Hao, Hong-Xing Yu, and Wei-Shi Zheng.
\newblock Mixgan: learning concepts from different domains for mixture
  generation.
\newblock In {\em IJCAI}, 2018.

\bibitem{isola2017image}
Phillip Isola, Jun-Yan Zhu, Tinghui Zhou, and Alexei~A Efros.
\newblock Image-to-image translation with conditional adversarial networks.
\newblock In {\em CVPR}, 2017.

\bibitem{jahanian2020steerability}
Ali Jahanian, Lucy Chai, and Phillip Isola.
\newblock On the''steerability" of generative adversarial networks.
\newblock In {\em ICLR}, 2020.

\bibitem{johnson2016perceptual}
Justin Johnson, Alexandre Alahi, and Li Fei-Fei.
\newblock Perceptual losses for real-time style transfer and super-resolution.
\newblock In {\em ECCV}, 2016.

\bibitem{pggan}
Tero Karras, Timo Aila, Samuli Laine, and Jaakko Lehtinen.
\newblock Progressive growing of gans for improved quality, stability, and
  variation.
\newblock In {\em ICLR}, 2018.

\bibitem{stylegan}
Tero Karras, Samuli Laine, and Timo Aila.
\newblock A style-based generator architecture for generative adversarial
  networks.
\newblock In {\em CVPR}, 2019.

\bibitem{kim2019grdn}
Dong-Wook Kim, Jae Ryun~Chung, and Seung-Won Jung.
\newblock Grdn: Grouped residual dense network for real image denoising and
  gan-based real-world noise modeling.
\newblock In {\em CVPR Workshop}, 2019.

\bibitem{glow}
Durk~P Kingma and Prafulla Dhariwal.
\newblock Glow: Generative flow with invertible 1x1 convolutions.
\newblock In {\em NeurIPS}, 2018.

\bibitem{lample2017fader}
Guillaume Lample, Neil Zeghidour, Nicolas Usunier, Antoine Bordes, Ludovic
  Denoyer, and Marc'Aurelio Ranzato.
\newblock Fader networks: Manipulating images by sliding attributes.
\newblock In {\em NeurIPS}, 2017.

\bibitem{ledig2017photo}
Christian Ledig, Lucas Theis, Ferenc Husz{\'a}r, Jose Caballero, Andrew
  Cunningham, Alejandro Acosta, Andrew Aitken, Alykhan Tejani, Johannes Totz,
  Zehan Wang, et~al.
\newblock Photo-realistic single image super-resolution using a generative
  adversarial network.
\newblock In {\em CVPR}, 2017.

\bibitem{liang2018generative}
Xiaodan Liang, Hao Zhang, Liang Lin, and Eric Xing.
\newblock Generative semantic manipulation with mask-contrasting gan.
\newblock In {\em ECCV}, 2018.

\bibitem{lipton2017precise}
Zachary~C Lipton and Subarna Tripathi.
\newblock Precise recovery of latent vectors from generative adversarial
  networks.
\newblock In {\em ICLR Workshop}, 2017.

\bibitem{liu2019few}
Ming-Yu Liu, Xun Huang, Arun Mallya, Tero Karras, Timo Aila, Jaakko Lehtinen,
  and Jan Kautz.
\newblock Few-shot unsupervised image-to-image translation.
\newblock In {\em ICCV}, 2019.

\bibitem{invertibility}
Fangchang Ma, Ulas Ayaz, and Sertac Karaman.
\newblock Invertibility of convolutional generative networks from partial
  measurements.
\newblock In {\em NeurIPS}, 2018.

\bibitem{niqe}
Anish Mittal, Rajiv Soundararajan, and Alan~C Bovik.
\newblock Making a “completely blind” image quality analyzer.
\newblock {\em IEEE Signal Processing Letters}, 2012.

\bibitem{inverting2016}
Guim Perarnau, Joost Van De~Weijer, Bogdan Raducanu, and Jose~M {\'A}lvarez.
\newblock Invertible conditional gans for image editing.
\newblock In {\em NeurIPS Workshop}, 2016.

\bibitem{shen2020interpreting}
Yujun Shen, Jinjin Gu, Xiaoou Tang, and Bolei Zhou.
\newblock Interpreting the latent space of gans for semantic face editing.
\newblock In {\em CVPR}, 2020.

\bibitem{shen2018faceid}
Yujun Shen, Ping Luo, Junjie Yan, Xiaogang Wang, and Xiaoou Tang.
\newblock Faceid-gan: Learning a symmetry three-player gan for
  identity-preserving face synthesis.
\newblock In {\em CVPR}, 2018.

\bibitem{vgg}
Karen Simonyan and Andrew Zisserman.
\newblock Very deep convolutional networks for large-scale image recognition.
\newblock In {\em ICLR}, 2015.

\bibitem{suarez2017infrared}
Patricia~L Su{\'a}rez, Angel~D Sappa, and Boris~X Vintimilla.
\newblock Infrared image colorization based on a triplet dcgan architecture.
\newblock In {\em CVPR Workshop}, 2017.

\bibitem{ulyanov2018deep}
Dmitry Ulyanov, Andrea Vedaldi, and Victor Lempitsky.
\newblock Deep image prior.
\newblock In {\em CVPR}, 2018.

\bibitem{upchurch2017deep}
Paul Upchurch, Jacob Gardner, Geoff Pleiss, Robert Pless, Noah Snavely, Kavita
  Bala, and Kilian Weinberger.
\newblock Deep feature interpolation for image content changes.
\newblock In {\em CVPR}, 2017.

\bibitem{wang2018high}
Ting-Chun Wang, Ming-Yu Liu, Jun-Yan Zhu, Andrew Tao, Jan Kautz, and Bryan
  Catanzaro.
\newblock High-resolution image synthesis and semantic manipulation with
  conditional gans.
\newblock In {\em CVPR}, 2018.

\bibitem{wang2018esrgan}
Xintao Wang, Ke Yu, Shixiang Wu, Jinjin Gu, Yihao Liu, Chao Dong, Yu Qiao, and
  Chen Change~Loy.
\newblock Esrgan: Enhanced super-resolution generative adversarial networks.
\newblock In {\em ECCV Workshop}, 2018.

\bibitem{ssim}
Zhou Wang, Alan~C Bovik, Hamid~R Sheikh, Eero~P Simoncelli, et~al.
\newblock Image quality assessment: from error visibility to structural
  similarity.
\newblock {\em TIP}, 2004.

\bibitem{yang2019semantic}
Ceyuan Yang, Yujun Shen, and Bolei Zhou.
\newblock Semantic hierarchy emerges in deep generative representations for
  scene synthesis.
\newblock {\em arXiv preprint arXiv:1911.09267}, 2019.

\bibitem{yeh2017semantic}
Raymond~A Yeh, Chen Chen, Teck Yian~Lim, Alexander~G Schwing, Mark
  Hasegawa-Johnson, and Minh~N Do.
\newblock Semantic image inpainting with deep generative models.
\newblock In {\em CVPR}, 2017.

\bibitem{lsun}
Fisher Yu, Ari Seff, Yinda Zhang, Shuran Song, Thomas Funkhouser, and Jianxiong
  Xiao.
\newblock Lsun: Construction of a large-scale image dataset using deep learning
  with humans in the loop.
\newblock {\em arXiv preprint arXiv:1506.03365}, 2015.

\bibitem{yu2018generative}
Jiahui Yu, Zhe Lin, Jimei Yang, Xiaohui Shen, Xin Lu, and Thomas~S Huang.
\newblock Generative image inpainting with contextual attention.
\newblock In {\em CVPR}, 2018.

\bibitem{zhang2016colorful}
Richard Zhang, Phillip Isola, and Alexei~A Efros.
\newblock Colorful image colorization.
\newblock In {\em ECCV}, 2016.

\bibitem{zhang2018unreasonable}
Richard Zhang, Phillip Isola, Alexei~A Efros, Eli Shechtman, and Oliver Wang.
\newblock The unreasonable effectiveness of deep features as a perceptual
  metric.
\newblock In {\em CVPR}, 2018.

\bibitem{rcan}
Yulun Zhang, Kunpeng Li, Kai Li, Lichen Wang, Bineng Zhong, and Yun Fu.
\newblock Image super-resolution using very deep residual channel attention
  networks.
\newblock In {\em ECCV}, 2018.

\bibitem{zhou2017scene}
Bolei Zhou, Hang Zhao, Xavier Puig, Sanja Fidler, Adela Barriuso, and Antonio
  Torralba.
\newblock Scene parsing through ade20k dataset.
\newblock In {\em CVPR}, 2017.

\bibitem{zhu2016generative}
Jun-Yan Zhu, Philipp Kr{\"a}henb{\"u}hl, Eli Shechtman, and Alexei~A Efros.
\newblock Generative visual manipulation on the natural image manifold.
\newblock In {\em ECCV}, 2016.

\bibitem{zhu2017unpaired}
Jun-Yan Zhu, Taesung Park, Phillip Isola, and Alexei~A Efros.
\newblock Unpaired image-to-image translation using cycle-consistent
  adversarial networks.
\newblock In {\em ICCV}, 2017.

\end{thebibliography}
}

\end{document}